\documentclass{IEEEcsmag}

\usepackage[colorlinks,urlcolor=blue,linkcolor=blue,citecolor=blue]{hyperref}
\expandafter\def\expandafter\UrlBreaks\expandafter{\UrlBreaks\do\/\do\*\do\-\do\~\do\'\do\"\do\-}
\usepackage{upmath,color}

\usepackage{graphicx}
\usepackage{eso-pic} 
\usepackage{subfig} 
\usepackage{transparent}
\usepackage{caption}
\usepackage{comment}
\usepackage{amsmath}
\usepackage{amsthm}
\usepackage{amssymb}
\usepackage{bm}
\usepackage[overload]{empheq}  

\usepackage{tabularx}
\usepackage{longtable} 
\usepackage{colortbl}
\usepackage{makecell} 
\usepackage{hhline}

\jvol{XX}
\jnum{XX}
\paper{8}
\jmonth{July}
\jname{IEEE Internet Computing}
\jtitle{IEEE Internet Computing}
\pubyear{2025}

\definecolor{red}{rgb}{0.76, 0.23, 0.13}
\definecolor{green}{rgb}{0.0, 0.42, 0.24}
\definecolor{yellow}{rgb}{1.0, 0.5, 0.0}
\definecolor{brown}{rgb}{0.6, 0.46, 0.33}
\definecolor{cyan}{rgb}{1.0, 0.98, 0.94}
\definecolor{y}{rgb}{1.0, 1.0, 0.0}

\begin{document}

\sptitle{\textsc{FEATURE ARTICLE: LLMs FOR DATA PREPARATION}}



\title{Lost in the Pipeline: How Well Do Large Language Models Handle Data Preparation?}

\author{Matteo Spreafico, Ludovica Tassini, Camilla Sancricca and Cinzia Cappiello}
\affil{Politecnico di Milano, Milan, Italy}

\markboth{LLMs for data preparation}{LLMs for data preparation}

\begin{abstract}
Large language models have recently demonstrated their exceptional capabilities in supporting and automating various tasks. Among the tasks worth exploring for testing large language model capabilities, we considered data preparation, a critical yet often labor-intensive step in data-driven processes. This paper investigates whether large language models can effectively support users in selecting and automating data preparation tasks. 
To this aim, we considered both general-purpose and fine-tuned tabular large language models. We prompted these models with poor-quality datasets and measured their ability to perform tasks such as data profiling and cleaning. We also compare the support provided by large language models with that offered by traditional data preparation tools. To evaluate the capabilities of large language models, we developed a custom-designed quality model that has been validated through a user study to gain insights into practitioners’ expectations. 
  
\end{abstract}


\maketitle

\chapteri{L}arge Language Models (LLMs) have exhibited remarkable capabilities across various domains, such as natural language understanding and problem-solving. They have revolutionized the field of artificial intelligence, demonstrating their ability to automate several data science tasks, such as producing reliable code or suggesting possible analyses \cite{Fang2024}.

More recently, fine-tuned models have been specifically designed for structured data (so-called ``tabular LLMs''), presenting a promising opportunity to support analytical processes with the most common type of data used in data analytics: tabular datasets.
However, whereas most of the literature studies that evaluate LLMs’ applications focus on unstructured data (text or images), testing their employment with tabular data remains largely unexplored.

Considering the growing importance of data-centric artificial intelligence, which emphasizes the critical role of data quality and data preparation in achieving reliable results in data analysis \cite{Jarrahi2023}, it is worth evaluating the potential of LLMs in supporting and automating the phases of a data analysis pipeline.

The most demanding phase of such a pipeline is data preparation, which includes tasks such as data profiling, transformation, and cleaning \--- a time-consuming process that can account for up to approximately 80\% of a data scientist's total time \cite{Hameed2020}.
In fact, data scientists often face challenges in creating an effective preparation pipeline that addresses a variety of quality errors and a multitude of available data preparation techniques.

This paper aims to investigate the effectiveness of LLMs in facilitating the preparation of tabular data. It discusses their strengths and limitations, providing guidelines on their practical utility and comparing their capacities with those of traditional data preparation tools.

First, to test the capabilities of LLMs, we designed a customized quality model that combines traditional data quality metrics, such as completeness and accuracy, with newly introduced metrics that represent relevant aspects to consider when evaluating LLMs' answers in this context.
To validate the proposed model, we conducted a user study involving 61 participants with diverse skills and expertise. We analyzed practitioners’ feedback to understand their needs and expectations regarding the adoption of LLMs to support the design of effective pipelines and refined the quality model with the gathered insights.

Second, we prompted LLMs to solve a set of data preparation tasks using datasets containing previously injected errors. 

We tested both general-purpose LLMs (i.e., GPT-4, Claude, Gemini, Llama, DeepSeek) and specialized tabular LLMs (i.e., TableGPT2, TableLLM) to determine their effectiveness in automating the following data preparation tasks: data profiling, dependency discovery, data wrangling (e.g., renaming, splitting, merging columns), and data cleaning. We also analyze data cleaning subtasks individually, such as data standardization (e.g., reorganization of composed fields, data type checks, or replacement of alternative spellings with a single one), data imputation, outlier detection, and data deduplication.
Each metric was linked to a list of statements (i.e., a checklist), which we manually checked to assess the quality of the answers.

Results reveal significant differences between general-purpose and tabular-specific models. Some models show promising capabilities in automating certain aspects of data preparation, but their effectiveness varies depending on the task and dataset's characteristics.

The paper is structured as follows. Section \hyperref[sec:related]{2} shows existing literature on data preparation tools, LLMs application with tabular data, and available techniques for evaluating LLMs; Section \hyperref[sec:pipeline]{3} presents the experimental pipeline and the design of our evaluation framework, which has been validated in Section \hyperref[sec:study]{4}; Section \hyperref[sec:results]{5} describes the obtained results and Section \hyperref[sec:conclusion]{6} concludes the paper and discusses future work.

\section{RELATED WORK}\label{sec:related}

\medskip

With the recent spread of data analysis, a growing number of approaches have been developed to help and support users in the phases of a data science pipeline, such as data exploration, profiling \cite{Ehrlinger2022}, DQ assessment \cite{Shrivastava2019}, or wrangling \cite{Yang2021}.
However, the main focus of these research efforts was supporting data cleaning with different objectives, such as automating the whole process \cite{RekatsinasCIR17, Berti2019} or suggesting users promising pipelines for new datasets \cite{Mahdavi2021, Zagatti2021}.
Other relevant works survey several tools for data quality analysis and data preparation \cite{Hameed2020, Patel2022}.
Given the large amount of research efforts to facilitate data preparation, exploring the potential of LLMs in supporting this demanding task is becoming crucial.

Recently, some works investigated what LLMs can do while working with tabular data, such as prediction, data synthesis, question answering, table understanding \cite{Fang2024, Lu2025}, row/column retrieval, table partition, or cell lookup \cite{Sui2024}. However, there is a lack of approaches examining LLMs' capabilities in preparing data.

A recent work \cite{Li2024} leverages LLMs to generate data-cleaning workflows focusing on issues such as duplicates, missing values, and inconsistent formats. Given a raw dataset and a query, it produces a minimally cleaned table and the corresponding workflow. Results demonstrate that LLMs can generate purpose-driven workflows without fine-tuning, advancing automation in data preprocessing. However, this study only focuses on data cleaning and uses evaluation metrics that miss relevant aspects to consider in the answers (as we demonstrated in the user study of Section \hyperref[sec:study]{4}).

Beyond general-purpose LLMs for data cleaning, recent advancements have focused on tabular LLMs, specifically fine-tuned for structured data. TableGPT2 integrates a dedicated table encoder, improving its ability to handle ambiguous queries, missing column names, and irregular table formats. Benchmarking results show it outperforms previous models in table-related tasks, making it highly relevant for business intelligence \cite{Su2024}. 

Similarly, TableLLM is optimized for tabular data manipulation, processing document-based and spreadsheet-embedded tables. It employs a hybrid approach, combining language modeling for documents with code-driven execution for spreadsheets. TableLLM demonstrates strong performance, surpassing GPT-3.5 and GPT-4 in spreadsheet-related tasks, marking a significant step toward LLMs that handle structured, real-world data \cite{Zhang2024}. In this paper, we aim to determine whether fine-tuning LLMs improves their performance in tabular data preparation.

Finally, a relevant set of contributions aims to evaluate LLMs' answers with LLMs evaluators, i.e., using LLMs themselves to assess their own outputs \cite{GPTScore, Shankar2024, Shankar2024SPADE}. While human evaluation is labor-intensive, LLM-based evaluation approaches have inherent limitations, as they may fail to capture nuanced aspects of language understanding; moreover, LLM-based evaluators inherit the same biases and limitations as the models they assess, making additional human validation necessary. For this reason, in this paper we adopted a human evaluation approach.

\section{EXPLORATORY WORKFLOW}\label{sec:pipeline}

\medskip

\begin{figure}[ht]
    \centering
    \includegraphics[width=0.49\textwidth]{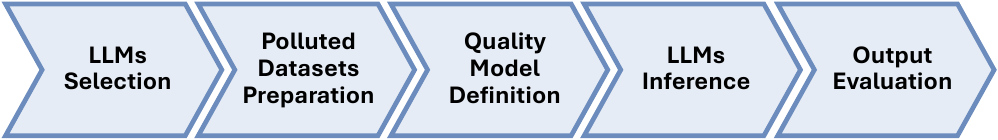}
    \caption{Exploratory workflow}
    \label{fig:pipeline}
\end{figure}

We implemented an experimental workflow to systematically test and evaluate the selected LLMs in supporting data preparation. Figure \ref{fig:pipeline} reports the main steps; they are detailed as follows.

\textbf{LLMs Selection} We selected the following General-purpose LLMs (GLLMs):  
\begin{itemize}
    \item \textit{Claude} (CLA): \texttt{claude-3-5-sonnet}
    \item \textit{Gemini} (GEM): \texttt{gemini-1.5-pro}
    \item \textit{GPT} (GPT): \texttt{gpt-4o-2024-08-06}
    \item \textit{Llama} (LLA): \texttt{llama-3.3-70b-versatile}
    \item \textit{DeepSeek} (DEEP): \texttt{DeepSeek-V3-0324}
\end{itemize}  

GLLM outputs were obtained via API calls to server endpoints.

\medskip

The selected Tabular LLMs (TLLMs) are as follows:  
\begin{itemize}
    \item \textit{TableGPT2} (tGPT): \texttt{tablegpt/TableGPT2-7B}
    \item \textit{TableLLM} (tLLM): \texttt{RUCKBReasoning/ TableLLM-13b}
\end{itemize}  

TLLM outputs were obtained by executing the models on a virtual machine. \textit{TableLlama} (\texttt{osunlp/TableLlama}) was also considered in the study, but it has been excluded from the evaluation due to poor textual output quality.

\textbf{Polluted Datasets Preparation} We work with a 100-row sample from a dataset taken from the Kaggle repository\footnote{\url{https://www.kaggle.com/datasets/ahmedshahriarsakib/usa-real-estate-dataset}}, containing transactional data on real estate in the USA. This dataset was chosen due to the variety of its column types (categorical, numerical, and date-based) and the absence of unstructured content (e.g., text). 

From the original dataset, we generated several polluted versions (with 10\%, 30\%, and 50\% percentage of injected errors), modifying the data using task-specific pollution functions: 
\begin{enumerate}
    \item \textit{Outliers injection} (used to evaluate outlier detection) inserts outliers in the numerical columns in order to simulate anomalies.
    
    \item \textit{De-standardization injection} (used to evaluate data standardization) introduces format inconsistencies in specifically selected columns (avoiding polluting primary key columns).
    
    \item \textit{Missing data injection} (used to evaluate data imputation) inserts missing values in numerical and categorical columns using different representations, such as empty strings, \texttt{nan}, \texttt{-1}, or \texttt{Unknown}. This allows for evaluating the ability of LLMs to identify both explicit and non-explicit missing values.
    
    \item \textit{Duplicate injection} (used to evaluate data deduplication) introduces exact and non-exact duplicates by replacing existing rows. Non-exact duplicates include manually introduced typos.
    
    \item \textit{Structural issues injection} (used to evaluate data wrangling) inserts suboptimal structural changes, such as column concatenation, column splitting, redundant column addition, and inconsistent naming conventions. For this pollution function, we generated a single polluted version.
    
    \item \textit{Dependency injection} (used to evaluate dependency discovery) modifies data to inject both relaxed and non-relaxed functional dependencies, aiming to test the ability of LLMs to detect them. For this pollution function, we generated a single polluted version.
\end{enumerate}

In order to evaluate the data profiling and cleaning task, we applied 1), 2), 3), and 4) pollution functions. The pollution process begins with a modified version of the duplicate injection, followed by a mask-based approach to selectively inject a single type of error in each targeted position of the table, avoiding the application of multiple pollution functions to the same value.

\textbf{Quality Model Definition} To evaluate LLM textual outputs, we designed a quality model that includes both traditional (i.e., accuracy and completeness) and new aspects (i.e., prescriptivity, readiness, and specificity) that can be relevant for a user who wants to be supported by an LLM in performing data preparation. Each task is associated with its own set of metrics among the following:

\begin{itemize}
    \item \textit{Completeness} ({COM}) is the degree to which all the pieces of information necessary to fully solve the problem are present in the LLM’s output \cite{Batini2006}.
    \item \textit{Accuracy} (ACC) is the degree to which the LLM’s output contains information that is aligned with the ground truth \cite{Batini2006}.
    \item \textit{Prescriptivity} (PRE) is the degree to which the LLM’s output provides users with a clear and unambiguous path to follow for solving a problem.
    \item \textit{Readiness} (REA) is the degree to which the LLM’s output is immediately usable, meaning no further processing is required to address the user’s problem.
    \item \textit{Specificity} (SPE) is the degree to which the LLM’s output specifically addresses each column of a tabular dataset rather than aggregated subsets of columns or the entire dataset.
\end{itemize}

Certain metrics have been excluded from the evaluation based on task-specific peculiarities. The list of metrics considered for each task is shown in Table \ref{table:qualitymodel}. 
Data profiling does not require the evaluation of \textit{Specificity}, as it already operates at the column level. Additionally, since profiling is a descriptive task aimed at summarizing dataset characteristics rather than providing actionable cleaning operations, \textit{Prescriptivity} is irrelevant.
Data standardization does not require \textit{Specificity}  evaluation since the de-standardization injection function introduced different formatting inconsistencies across columns, and there is no non-specific solution applicable to this task.
For data deduplication, wrangling, and dependency discovery, \textit{Specificity} is not assessed since these tasks must be applied to the whole dataset rather than on specific columns.

\begin{table}
    \centering
    \includegraphics[width=\columnwidth]{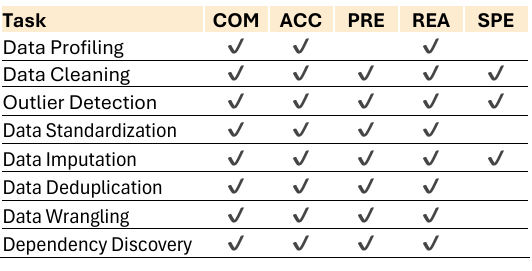}
    \caption{Metrics mapping with tasks}
    \label{table:qualitymodel}
\end{table}

\textbf{LLMs Inference} All LLMs are configured with a \textit{temperature} equal to 0 and fed with the same prompt:

\begin{verbatim}
    Consider this dataset:\n
    {{csv_text}}\n
    Can you do {{task_name}} on it?
\end{verbatim}

where:  

\begin{itemize}
    \item \texttt{\{\{csv\_text\}\}} is replaced with a textual version of the dataset in \textit{.csv} format;
    \item \texttt{\{\{task\_name\}\}} is replaced with the name of the task in analysis.
\end{itemize}

\textbf{Output Evaluation} LLM outputs are collected and then manually evaluated using checklists composed of a set of statements. Each metric has at least one specific checklist, although additional versions may exist to accommodate the varying levels of pollution. Each checklist includes a series of factual statements for which we must verify whether the given output aligns with the reported fact. The statements are evaluated with a score from 0 to 1; they can be binary (0 = false, 1 = true) or continuous (based on the specific fact).
For example, a fact belonging to the data profiling checklist for assessing \textit{Accuracy} is as follows: ``the LLM's response explicitly says that the data type of the \texttt{status} column is categoric''.

It is worth noting that \textit{Completeness} influences the assessment of the other metrics. In fact, verifying a fact from different perspectives (i.e., using different metrics) is significant only if the fact is present in the answer \--- if it has a \textit{Completeness} score greater than zero. Therefore, facts that are not present in the LLM output are marked as ``non-evaluable'' (N/A) in the other metrics' checklist. For example, in the data imputation checklist, one statement used to evaluate the \textit{Completeness} of an answer is: ``A solution for imputing \texttt{NaN} values for the \texttt{bed} column is proposed''; if this statement has a zero \textit{Completeness} score, then, all the other statements referring to this fact are ``non-evaluable''.

Metrics are calculated as the ratio of true facts to the total number of evaluable facts. However, some checklists employ a weighted evaluation of the statements where specific facts have greater significance:
    \begin{itemize}
        \item In \textit{Accuracy} checklists such as data standardization and imputation, the statements verify that the LLM's proposed solutions are \emph{valid} (i.e., they can be applied, but they are not the optimal tasks, for example, applying standard imputation to all the columns) or \emph{optimal} (for example applying advanced/ad-hoc imputation techniques to different columns). The facts associated with a valid solution contribute to 80\% of the total score, while the optimal solution statements account for the remaining 20\%.
        \item In all \textit{Readiness} checklists, the statements verify if the LLM provides ``code'' or ``ready-to-use data''. The ``code'' statements contribute 80\% of the total score, while ``ready-to-use data'' statements account for the remaining 20\%.
    \end{itemize}
    
Certain \textit{Accuracy} checklists (for outlier detection, data deduplication, and dependency discovery), rather than with a binary score, categorize True Positives, False Positives, and False Negatives to quantify how many outliers, duplicates, and functional dependencies were correctly detected, misclassified or not identified. In these cases, \textit{Accuracy} is calculated using the F1 score derived from precision and recall.

Finally, \textit{Specificity}, rather than with a binary score, is evaluated by measuring how many columns the LLM proposes ad-hoc solutions for a specific statement (for example, ``A solution for imputing \texttt{NaN} values is proposed for column C''). The overall \textit{Specificity} score is obtained by averaging these values for all statements.


\section{USER STUDY}\label{sec:study}
\medskip

To validate the assumptions behind the newly introduced quality metrics, we conducted a user study with 61 international participants, including 41 experts and 20 non-experts. Experts are data practitioners, with 51\% identifying as data scientists or data engineers, while the remaining 49\% have different working positions.
Non-experts have no professional experience working with data; however, 95\% of the non-experts reported having background knowledge in data preparation.


The study consisted of a questionnaire where we asked users, given a set of LLMs' answers as an example, to evaluate on a liking scale their agreement or perceived support concerning a set of statements. The questionnaire's results are accessible in a public repository\footnote{\url{https://github.com/MattBlue00/polimi-thesis/tree/main/user_study_results}}.

Specifically, for \textit{Prescriptivity}, we aimed to verify whether users prefer prescriptive or non-prescriptive answers. For \textit{Readiness}, we evaluated whether structured numerical data, code, and textual suggestions are appropriately ranked in terms of their usefulness for supporting data preparation. Finally, for \textit{Specificity}, we examined whether users favor specific over general answers and whether they accept variability in the granularity (i.e., column-specific or general solutions) of responses. The main findings of the user study are detailed as follows.

\paragraph{Metrics' Importance}
Users consistently value \textit{Completeness} and \textit{Accuracy} as key factors in evaluating responses. However, \textit{Readiness} revealed a difference in the opinions of users with different expertise: more experienced users prioritize ready-to-use outputs, while less experienced users prefer more guidance in completing the tasks.
\textit{Prescriptivity} is more relevant for less experienced users, who prefer structured guidance, whereas more experienced users favor more flexibility in trying different tasks. Similarly, \textit{Specificity} is valued by all users, but those with more experience demonstrate greater tolerance between column-specific and generalized solutions, while those less experienced favor very specific solutions. 

\paragraph{Desired Level of Support} One aspect of the study focused on whether users favor ready-to-use support, such as code or numerical data, over textual suggestions. Numerical data emerged as the most preferred format, yielding the highest satisfaction levels. Code-based support received mixed reactions, with opinions varying based on users’ familiarity with coding. Finally, text-based descriptions were moderately appreciated but generally considered less useful than structured data.

Another objective was to determine whether users prefer prescriptive support, where a single predefined strategy is provided for task completion, over multiple alternatives, which then requires making decisions on what to do individually. Users tend to favor a single predefined strategy, but experience plays a role: less experienced users seek structured guidance, whereas experts prefer more flexibility in their choices.

The study also investigated whether users prefer specific, column-oriented support over a more general, table-wise approach. Column-oriented solutions are strongly favored over the general table-wise approach. More experienced users exhibit slightly greater flexibility, but the majority of users favor fine-grained, column-specific support. The aggregated approach was slightly less preferred but still received positive feedback.

\paragraph{Additional Feedback} At the end of the questionnaire, we put an open-ended answer to let users provide additional feedback. Open-ended feedback highlights the need for clearer explanations and the importance of balancing automation with user decisions, particularly in cases involving code generation, where explanations alongside the code are considered essential.

\paragraph{Model Refinements}
The user study confirmed that all the considered metrics of the proposed quality model are relevant to users. \textit{Completeness}, \textit{Accuracy}, and \textit{Specificity} require no refinements.
For \textit{Prescriptivity}, while different perspectives have been provided based on the user's expertise, we envision that users will interpret the score according to their experience and needs; thus, we do not change the model or the scoring function.
For \textit{Readiness}, we performed the following refinements:
\begin{itemize}
    \item Code suggestions were initially assigned to a higher score but received mixed feedback due to a lack of explanations; thus, the score is lowered from 0.8/1 to 0.5/1.
    \item Returning a cleaned dataset was initially scored 1/1, but users preferred receiving only the set of tuples involved in the task; therefore, the score should be reduced to 0.8/1. However, returning only a subset of data is appropriate only for data deduplication while for tasks such as data imputation, this approach is impractical. This modification is implemented only for the data deduplication task/data cleaning subtask.
\end{itemize}
Textual suggestions were initially scored with 0 \textit{Readiness}, but contrary to our expectations, they were perceived by users as comparable to code. However, user feedback indicates that dissatisfaction arises from a lack of explanations rather than from an insufficient ``readiness''; thus, this score remains unchanged.
These refinements align the scoring mechanisms with user preferences while preserving the quality model.

\section{EVALUATION RESULTS}\label{sec:results}

\medskip

This section presents the results obtained by executing the pipeline described in Section \hyperref[sec:pipeline]{3}. The LLMs' outputs were evaluated using the final quality model, refined after the user study.
For each task/dataset/LLM combination, an output is generated and evaluated using the checklists described in Section \hyperref[sec:pipeline]{3}. The aggregated results for each task are presented in Table \ref{table:results} and discussed as follows, while the complete set of evaluations is available in a public repository\footnote{\url{https://github.com/MattBlue00/polimi-thesis}}.

\begin{table*}
    \centering
    \includegraphics[width=\textwidth]{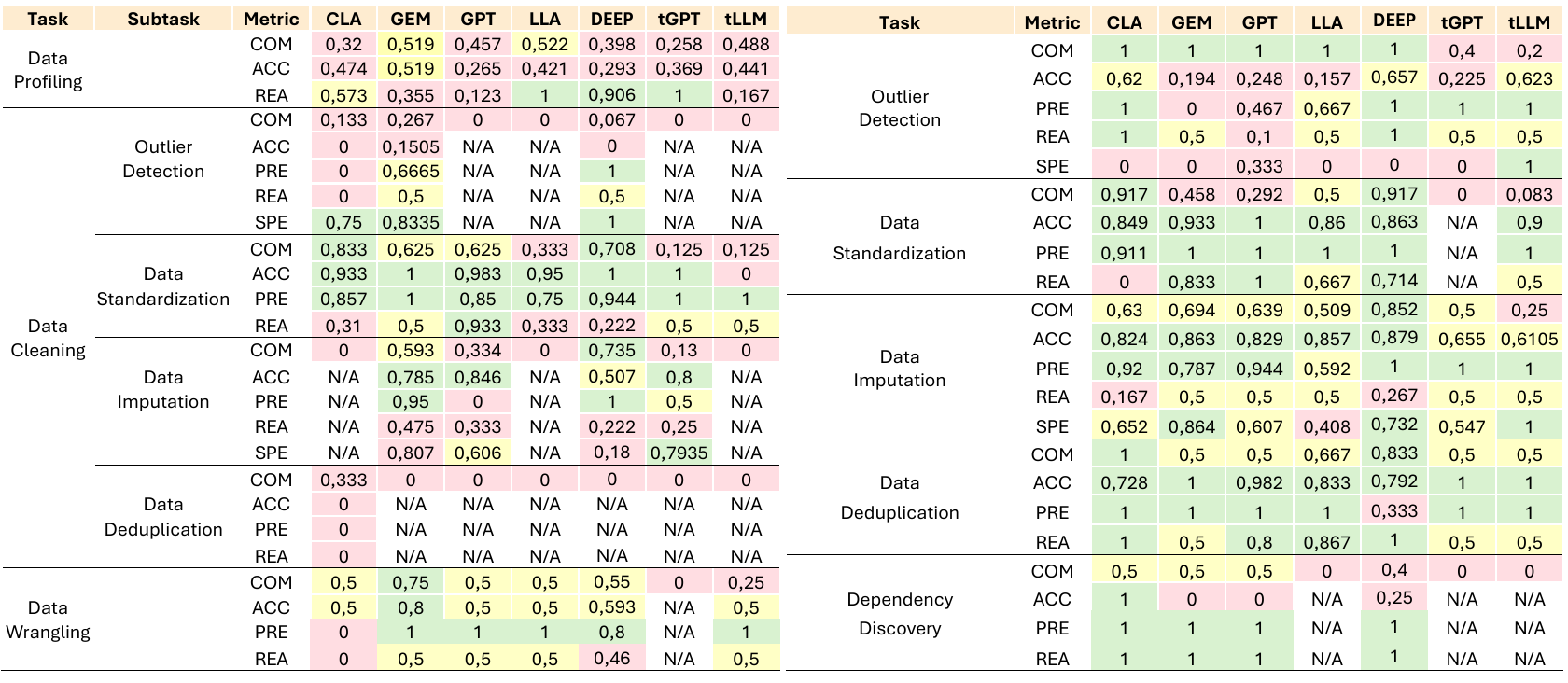}
    \caption{Evaluation results}
    \label{table:results}
\end{table*}

\paragraph{Data Profiling} GLLMs perform only basic profiling actions, leaving more complex subtasks \--- when mentioned \--- as mere suggestions. In general, they achieve a medium \textit{Completeness} score, meaning that half of the expected profiling tasks were reported in the answers. \textit{Accuracy} yields worse results due to the well-known limitations of LLMs in counting tasks. All GLLMs except Llama and DeepSeek fail to provide ready-to-use data/information. 

TableLLM achieves the best performance among TLLMs. However, upon examining its output, we find that it merely suggests using a (deprecated) profiling tool from the \texttt{pandas} library, rendering the use of an LLM unnecessary.

\paragraph{Data Cleaning} Data cleaning is complex and has multiple subtasks, and GLLMs execute only a very limited subset of them. Among the subtasks we consider, they mainly perform data standardization and, less frequently, data imputation. Gemini and DeepSeek stand out as the only models that also attempt outlier detection without completely failing. The data deduplication task is always skipped. Considering data cleaning, we generally observe a degradation of LLM performance as data dirtiness increases.

TLLMs provide limited support for data standardization and data imputation, and no support at all for the other two subtasks. We note that TableLLM often fails entirely, not including any relevant content in its answers. TableGPT2's answers are very incomplete, but at least they are more similar to those provided by GLLMs.

\paragraph{Outlier Detection} GLLMs are capable of performing outlier detection tasks, although their performance significantly degrades as dataset dirtiness (i.e., number of injected outliers) increases. Claude and DeepSeek stand out as the most effective GLLM, consistently delivering accurate, prescriptive, and ready-to-use results.  

TLLMs exhibit significantly lower \textit{Completeness}, as they focus on at most two out of five numerical columns (with outliers). While TableLLM’s overall performance is acceptable, aside from \textit{Completeness}, TableGPT2 shows a decline in performance as the number of outliers increases: it detects half of them or even none.

\paragraph{Data Standardization} All LLMs show consistently excellent \textit{Accuracy} for this task, while \textit{Completeness} is slightly lower. GPT often performs data normalization (e.g., scaling) instead of data standardization, resulting in lower \textit{Completeness}. While responses tend to be highly prescriptive, \textit{Readiness} varies significantly across GLLMs.

TLLMs show poor data standardization capabilities: TableGPT2 confuses data standardization with data normalization, while TableLLM correctly addresses only a few of the injected standardization issues.

\paragraph{Data Imputation} GLLMs demonstrate significant capabilities in data imputation. All of them have similar \textit{Completeness} and \textit{Accuracy} scores, even across datasets with different levels of dirtiness (i.e., varying numbers of missing values). While all GLLMs effectively detect missing values in their standard format (e.g., empty strings), they struggle with more complex formats (e.g., \texttt{-1} or special characters as \texttt{\---}). Generally, the imputation methods they suggest are appropriate, but they rarely consider advanced techniques (e.g., ML-based approaches).  

TLLMs provide inadequate support for data imputation, achieving medium \textit{Completeness} and \textit{Accuracy} scores. Moreover, TLLMs detect only trivial missing values (e.g., empty strings).

\paragraph{Data Deduplication} GLLMs demonstrate reliable and optimal performance in detecting exact duplicates. However, their performance declines when handling non-exact duplicates, which are almost never detected. 
\textit{Prescriptivity} is excellent except for DeepSeek, and \textit{Readiness} maintains almost always a high score, meaning that they provide prescriptive answers and ready-to-use data.

TLLMs never account for non-exact duplicates. While all metrics \--- except \textit{Completeness} and \textit{Readiness} \--- are excellent, the lack of recognizing non-exact duplicates highlights LLMs' uselessness in supporting this task, as exact duplicates can be trivially retrieved using traditional tools (e.g., \texttt{pandas}).

\paragraph{Data Wrangling} GLLMs prove to support data wrangling effectively, suggesting actions such as splitting and merging columns, and sometimes recommending the removal of unnecessary ones. While most GLLMs execute merge and split operations correctly, we notice that their suggestions are not always the optimal ones.

TLLMs demonstrate a limited capacity of handling data-wrangling tasks. While TableGPT2 mainly suggests tasks related to data cleaning, TableLLM shows a slightly better performance by addressing at least one of the four structural issues introduced by the pollution function (i.e., column concatenation, column splitting, redundant column addition, and inconsistent naming conventions for broker information).

\paragraph{Dependency Discovery} Neither GLLMs nor TLLMs are capable of performing this task, sometimes even explicitly stating their inability to perform it. Often, they refer to correlation analysis as a substitute, which, while potentially helpful in identifying dependencies, is theoretically different from dependency discovery. For this reason, LLMs cannot replace traditional tools in performing this task.

\paragraph{Best LLMs}
Table \ref{table:bestllms} indicates the best-performing LLMs for each task. In this way, users can leverage the capabilities of each model to support their tabular data preparation pipelines. We highlight that GLLMs always outperform TLLMs in every task.

\begin{table}
    \centering
    \includegraphics[width=0.6\columnwidth]{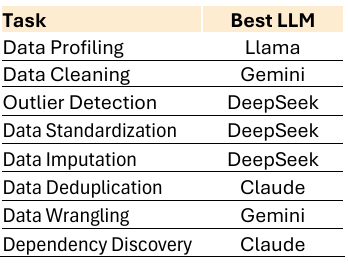}
    \caption{Best-performing LLMs for each task}
    \label{table:bestllms}
\end{table}

\section{CONCLUSIONS}\label{sec:conclusion}
\medskip
This paper is an initial step toward the understanding of LLMs' capabilities in supporting data preparation of tabular data.
We found that LLMs can assist users in performing data preparation tasks with varying effectiveness, depending on the task and the dataset's level of dirtiness.
It is worth noting the current limitations of LLMs, such as issues with counting tasks and fixed input length, while traditional preparation algorithms effectively perform counting tasks and have unlimited input capabilities. 
These relevant factors strengthen that \textbf{LLMs cannot yet replace traditional data preparation tools}. However, they can support those tools by leveraging their context-understanding capabilities to provide insights that algorithm-based tools cannot offer.
Also, we found that current \textbf{general-purpose LLMs clearly outperform fine-tuned tabular LLMs}. 

Future research directions can include (i) leveraging the interaction with LLMs and iterative refinement (i.e., prompt engineering) to improve data preparation support, (ii) automating the evaluation framework instead of manually assessing the answers, and (iii) extending the evaluation across diverse datasets and a broader range of tasks.

\bibliographystyle{IEEEtran}
\bibliography{bibliography.bib}

@inproceedings{Zagatti2021,
  author       = {Fernando Rezende Zagatti and
                  Lucas Cardoso Silva and
                  Lucas Nildaimon dos Santos Silva and
                  Bruno Silva Sette and
                  Helena de Medeiros Caseli and
                  Daniel Lucr{\'{e}}dio and
                  Diego Furtado Silva},
  title        = {MetaPrep: Data preparation pipelines recommendation via meta-learning},
  booktitle    = {{ICMLA}},
  pages        = {1197--1202},
  publisher    = {{IEEE}},
  year         = {2021}
}

@article{Jarrahi2023,
  author       = {Mohammad Hossein Jarrahi and
                  Ali Memariani and
                  Shion Guha},
  title        = {The Principles of Data-Centric {AI}},
  journal      = {Commun. {ACM}},
  volume       = {66},
  number       = {8},
  pages        = {84--92},
  year         = {2023}
}

@article{Hameed2020,
  author       = {Mazhar Hameed and
                  Felix Naumann},
  title        = {Data Preparation: {A} Survey of Commercial Tools},
  journal      = {{SIGMOD} Rec.},
  volume       = {49},
  number       = {3},
  pages        = {18--29},
  year         = {2020}
}

@inproceedings{Shrivastava2019,
  author       = {Shrey Shrivastava and
                 others},
  title        = {{DQA:} Scalable, Automated and Interactive Data Quality Advisor},
  booktitle    = {Proc. of 2019 {(IEEE} BigData)},
  pages        = {2913--2922},
  publisher    = {{IEEE}}
}

@inproceedings{Berti2019,
  author       = {Laure Berti{-}{\'{E}}quille},
  title        = {Learn2Clean: Optimizing the Sequence of Tasks for Web Data Preparation},
  booktitle    = {The World Wide Web Conference, {WWW} 2019},
  pages        = {2580--2586},
  publisher    = {{ACM}}
}

@inproceedings{Mahdavi2021,
  author       = {Mohammad Mahdavi and
                  Ziawasch Abedjan},
  title        = {Semi-Supervised Data Cleaning with Raha and Baran},
  booktitle    = {11th Conference on Innovative Data Systems Research, {CIDR} 2021},
  publisher    = {www.cidrdb.org}
}

@article{RekatsinasCIR17,
  author       = {Theodoros Rekatsinas and
                  Xu Chu and
                  Ihab F. Ilyas and
                  Christopher R{\'{e}}},
  title        = {HoloClean: Holistic Data Repairs with Probabilistic Inference},
  journal      = {Proc. {VLDB} Endow.},
  volume       = {10},
  number       = {11},
  pages        = {1190--1201},
  year         = {2017}
}

@book{Batini2006,
  author       = {Carlo Batini and
                  Monica Scannapieco},
  title        = {Data Quality: Concepts, Methodologies and Techniques},
  series       = {Data-Centric Systems and Applications},
  publisher    = {Springer},
  year         = {2006}
}

@article{Ehrlinger2022,
  author       = {Lisa Ehrlinger and
                  Wolfram W{\"{o}}{\ss}},
  title        = {A Survey of Data Quality Measurement and Monitoring Tools},
  journal      = {Frontiers Big Data},
  volume       = {5},
  pages        = {850611},
  year         = {2022}
}

@article{Yang2021,
  title={Auto-pipeline: synthesizing complex data pipelines by-target using reinforcement learning and search},
  author={Yang, Junwen and He, Yeye and Chaudhuri, Surajit},
  journal={PVLDB},
  volume = {14},
  number = {11},
  pages = {2563 -- 2575},
  year={2021}
}

@inproceedings{Patel2022,
  author       = {Hima Patel et al},
  title        = {Automatic Assessment of Quality of your Data for {AI}},
  booktitle    = {{CODS-COMAD} 2022: 5th Joint International Conference on Data Science
                  {\&} Management of Data (9th {ACM} {IKDD} {CODS} and 27th COMAD),
                  Bangalore, India, January 8 - 10, 2022},
  pages        = {354--357},
  publisher    = {{ACM}},
  year         = {2022}
}

@article{Li2024,
  author       = {Lan Li and
                  Liri Fang and
                  Vetle I. Torvik},
  title        = {AutoDCWorkflow: LLM-based Data Cleaning Workflow Auto-Generation and
                  Benchmark},
  journal      = {CoRR},
  volume       = {abs/2412.06724},
  year         = {2024}
}

@article{Su2024,
  author       = {Aofeng Su et al},
  title        = {TableGPT2: {A} Large Multimodal Model with Tabular Data Integration},
  journal      = {CoRR},
  volume       = {abs/2411.02059},
  year         = {2024}
}

@article{Zhang2024,
  author       = {Xiaokang Zhang et al},
  title        = {TableLLM: Enabling Tabular Data Manipulation by LLMs in Real Office
                  Usage Scenarios},
  journal      = {CoRR},
  volume       = {abs/2403.19318},
  year         = {2024}
}

@article{Fang2024,
  author       = {Xi Fang and
                  Weijie Xu and
                  Fiona Anting Tan and
                  Ziqing Hu and
                  Jiani Zhang and
                  Yanjun Qi and
                  Srinivasan H. Sengamedu and
                  Christos Faloutsos},
  title        = {Large Language Models (LLMs) on Tabular Data: Prediction, Generation,
                  and Understanding - {A} Survey},
  journal      = {Trans. Mach. Learn. Res.},
  volume       = {2024},
  year         = {2024}
}

@article{Lu2025,
  author       = {Weizheng Lu and
                  Jing Zhang and
                  Ju Fan and
                  Zihao Fu and
                  Yueguo Chen and
                  Xiaoyong Du},
  title        = {Large language model for table processing: a survey},
  journal      = {Frontiers Comput. Sci.},
  volume       = {19},
  number       = {2},
  pages        = {192350},
  year         = {2025}
}

@article{Shankar2024SPADE,
  author       = {Shreya Shankar and
                  Haotian Li and
                  Parth Asawa and
                  Madelon Hulsebos and
                  Yiming Lin and
                  J. D. Zamfirscu{-}Pereira and
                  Harrison Chase and
                  Will Fu{-}Hinthorn and
                  Aditya G. Parameswaran and
                  Eugene Wu},
  title        = {{SPADE:} Synthesizing Data Quality Assertions for Large Language Model
                  Pipelines},
  journal      = {Proc. {VLDB} Endow.},
  volume       = {17},
  number       = {12},
  pages        = {4173--4186},
  year         = {2024}
}

@inproceedings{Shankar2024,
  author       = {Shreya Shankar and
                  J. D. Zamfirescu{-}Pereira and
                  Bjoern Hartmann and
                  Aditya G. Parameswaran and
                  Ian Arawjo},
  title        = {Who Validates the Validators? Aligning LLM-Assisted Evaluation of
                  {LLM} Outputs with Human Preferences},
  booktitle    = {{UIST}},
  pages        = {131:1--131:14},
  publisher    = {{ACM}},
  year         = {2024}
}

@inproceedings{GPTScore,
  author       = {Jinlan Fu and
                  See{-}Kiong Ng and
                  Zhengbao Jiang and
                  Pengfei Liu},
  title        = {GPTScore: Evaluate as You Desire},
  booktitle    = {{NAACL-HLT}},
  pages        = {6556--6576},
  publisher    = {Association for Computational Linguistics},
  year         = {2024}
}

@inproceedings{Sui2024,
  author       = {Yuan Sui and
                  Mengyu Zhou and
                  Mingjie Zhou and
                  Shi Han and
                  Dongmei Zhang},
  title        = {Table Meets {LLM:} Can Large Language Models Understand Structured
                  Table Data? {A} Benchmark and Empirical Study},
  booktitle    = {{WSDM}},
  pages        = {645--654},
  publisher    = {{ACM}},
  year         = {2024}
}

@misc{claude,
  howpublished = "\url{https://www.anthropic.com/claude/sonnet}",
}

@misc{gemini,
  howpublished = "\url{https://blog.google/technology/ai/google-gemini-next-generation-model-february-2024/}",
}

@misc{gpt,
  howpublished = "\url{https://platform.openai.com/docs/models}",
}

@misc{llama,
  howpublished = "\url{https://www.llama.com/docs/model-cards-and-prompt-formats/llama3_3/}",
}

@misc{deepseek,
  howpublished = "\url{https://api-docs.deepseek.com/news/news250325}",
}

@misc{tablellm,
  howpublished = "\url{https://huggingface.co/RUCKBReasoning/TableLLM-13b}",
}

@misc{tablellama,
  howpublished = "\url{https://huggingface.co/osunlp/TableLlama}",
}


\end{document}